\documentclass[conference]{IEEEtran}
\IEEEoverridecommandlockouts
\usepackage{cite,flushend}
\usepackage{amsmath,amssymb,amsfonts}
\usepackage{algorithmic}
\usepackage{graphicx}
\usepackage{textcomp}
\usepackage{xcolor}
\usepackage{multirow}
\usepackage{commath}
\usepackage{fancyhdr,graphicx,amsmath,amssymb}
\usepackage[ruled,vlined]{algorithm2e}

\usepackage{fancyhdr}
\SetCommentSty{mycommfont}

\def\BibTeX{{\rm B\kern-.05em{\sc i\kern-.025em b}\kern-.08em
 T\kern-.1667em\lower.7ex\hbox{E}\kern-.125emX}}
\begin{document}

\title{\emph{Projectron} -- A Shallow and Interpretable Network for Classifying Medical Images\\
}

\author{\IEEEauthorblockN{Aditya Sriram$^1$, Shivam Kalra$^1$, H.R. Tizhoosh$^{1,2}$}
\IEEEauthorblockA{\textit{$^1$ Kimia Lab, University of Waterloo,
Ontario, Canada; kimia.uwaterloo.ca} \\
$^2$ Vector Institute, Toronto, Canada\\
asriram/s6kalra/tizhoosh@uwaterloo.ca}
}

\maketitle

\begin{abstract}
This paper introduces the ``Projectron'' as a new neural network architecture that uses Radon projections to both classify and represent medical images. The motivation is to build shallow networks which are more interpretable in the medical imaging domain. Radon transform is an established technique that can reconstruct images from parallel projections. The Projectron first applies global Radon transform to each image using equidistant angles and then feeds these transformations for encoding to a single layer of neurons followed by a layer of suitable kernels to facilitate a linear separation of projections. Finally, the Projectron provides the output of the encoding as an input to two more layers for final classification. 
We validate the Projectron on five publicly available datasets, a general dataset (namely MNIST) and four medical datasets (namely Emphysema, IDC, IRMA, and Pneumonia). The results are encouraging as we compared the Projectron's performance against MLPs with raw images and Radon projections as inputs, respectively. Experiments clearly demonstrate the potential of the proposed Projectron for representing/classifying medical images.
\end{abstract}


\section{Introduction}
Computer vision is a collection of techniques to, among others, extract features (also called descriptors) from images using handcrafted algorithms. The problem with such approaches is the limited scope as these algorithms are designed to only solve a certain task or application with no inherent capability of adjusting to the characteristics of new (unseen) images. Despite the large number of image descriptors that are available in literature, the principle of comparing features using some distance metrics or using them for classification  requires careful design and customized configuration. The leap forward is to develop techniques that learn/weigh these features automatically without any manual intervention. Over the years, Radon transform has gained some traction as an image descriptor in the medical imaging domain. The features captured by Radon transform are based on equi-distant angles that capture the shape of the objects and organs. So far, the usage of Radon projections as image features have been primarily based on developing handcrafted descriptors for retrieval approach  \cite{tizhoosh2015barcode} \cite{babaie2017local}\cite{tizhoosh2018representing} or  for classification \cite{zhu2016radon} \cite{sriram2017learning} \cite{khatami2018sequential}. These techniques have difficulty generalizing across different image instances for the same application. To overcome this shortcoming, we propose a new  network called ``Projectron'' that learns to represent/classify medical images using Radon projections as input. 

The proposed network is comprised of three phases: generating Radon projections, an encoding block, and a classification block (whose weights can also be used as representation). For every image, multiple Radon projections are calculated across equi-distant angles ranging between $0^{\circ}$ and $180^{\circ}$. 
To ensure that each Radon projection has the same length, the images are re-sized to have same-length width/height. As for the encoding block, the projections are passed onto a single-layer of neurons. This layer, considered isolated, is a binary classifier that decides whether the input, represented as a Radon projection vector, belong to a specific class. In this case, the first layer learns the Radon projection by combining a set of weights through ReLU activation function, to linearly classify all projections. A kernel layer follows the  first layer in order to transform the neurons' outputs into a more easily separable space. 
The last stage in the Projectron classification  is the two fully connected layers which intuitively carry the most dense features prior to the final reduction to the number of classes  through a traditional softmax classification scheme. 

The motivation for this research is twofold: 1) We would like to design ``shallow networks'' that are more easily trainable compared to deep architectures that require a lot of efforts for  design, implementation and training, and 2) we would like to work with networks in the medical domain whose results are more interpretable; when a decision is made by the network, it should be possible to ``understand'' that decision by observing/examining the input, a possibility that is not available when deep features are employed for classification.   

In this work, we introduce a new Radon-based neural network, called ``Projectron'', which learns and classifies Radon projections. Five public datasets (MNIST, Emphysema, IDC, IRMA, and Pneumonia) were adopted to evaluate the proposed network. The Projectron is compared against two conventional MLP networks - one with raw images as input, and the other where the inputs are the same Radon projections computed for the Projectron. The Projectron performs the same if not better than MLP in majority of the datasets. 

\section{Related Works}
Recently, Radon projections have gained some traction in computer vision as an image descriptor \cite{sanz2013radon}. One of the first versions of Radon projection as an image descriptor was proposed by Tizhoosh \cite{tizhoosh2015barcode}, called Radon Barcodes (RBC). The purpose was to binarize Radon projections to create a barcode annotation that is a short feature representation of an image. Validated on the IRMA dataset consisting of 14,410 radiograph images from 193 different classes, the RBC performed an IRMA error, \textcolor{black}{defined in Equation \ref{irma_error},} of 476.62 using only 4 projections forming a vector length of 512 digits. In 2016, Tizhoosh et al. \cite{tizhoosh2016minmax} introduced an improved RBC descriptor, called MinMax RBC. The authors apply a smoothing function before capturing the shape of projections. This enables the detection of all major minimum and maximum values in a projection profile, and allows for creating sections for more meaningful binarization of the projection. Subsequently, the authors assign an encoded value of 1 or 0 based on the slope of the projection profile between the sections. With the MinMax RBC, the IRMA error dropped to 415.75 and was observed to be become computationally much faster. A recent improvement of RBC was put forward by Babaie et al. \cite{babaie2017retrieving}, wherein the authors experiment if a single Radon projection can represent an image, called Single Projection Radon barcode (Sp-RBC). This study deduced that exploiting a single projection to form a short feature vector does provide acceptable results. To make Sp-RBC more robust, the authors use the outcome of each projection separately, and deduce the best-match using local search. Tested on the IRMA dataset, Sp-RBC yields an IRMA error of 356.57. Later, Babaie et al.~\cite{babaie2017local} explored the difference when capturing Radon features at a global and local neighborhoods of images. This strategy is particularly useful when the dimension of the image is too large to process, such as in digital pathology (i.e., whole slide imaging or WSI) and satellite images. The authors developed a descriptor called Local Radon Descriptor (LRD) that generates a histogram based on Radon projections computed at a local neighborhood across the entire image. This approach is observed to yield a higher discrimination of features as compared to global radon projection. The aforementioned descriptor is validated on IRMA dataset as well as INRIA holiday dataset (consisting of 1,990 images). LRD obtained an IRMA error of 287.77, and an accuracy of 40.02\% on the INRIA holiday dataset - which is a comparative score against well-established descriptors such as LBP and HoG.

A quasi-learning approach using Radon projections was investigated by Zhu and Tizhoosh \cite{zhu2016radon}. The authors used normalized Radon projections for extracting features from raw images, and provided these descriptors as an input to Support Vector Machines (SVM) for classification. In particular, the Radon features are binarized to form Radon barcodes which are then used to tag all images. For retrieving similar images, Radon barcodes are extracted for the query image and a k-nearest neighbor search is applied to find the best match using the minimum Hamming distance. This approach is observed to correctly identify image classes for those that are mistakenly classified by SVM. Experimental results yield an IRMA error of 294.83. More recently, Sriram et al. \cite{sriram2017learning} performed a study to determine if the use of Radon projections is a better feature for learning algorithms to generalize. The authors performed a comparative study, between Radon projections, histogram of oriented gradients (HoG) and raw (unprocessed) pixels, to determine the best image descriptor for an autoencoder to compress an image with minimal loss. The proposed framework extracted the aforementioned features from images, compressed those features using a shallow autoencoder, and passed them onto a Multi-Layer Perceptron (MLP) for classification. The authors observed that Radon features as an input vector to an autoencoder provided the best result. Validated on the IRMA dataset, the proposed framework achieved an IRMA error of 313 (equivalent to 82\% accuracy), which outperforms all other autoencoding approaches.

Apart from using Radon transform for classification, these descriptors are also used for narrowing the search-space for better image retrieval. In 2017, Khatami et al. \cite{khatami2017deep} used a deep CNN to classify radiograph images to obtain a set of ``best predicted categories''. To further narrow the query, Radon transform was adopted for similarity-based search schemes, after obtaining the k-nearest neighbors. This approach is observed to be fast as well as providing improved performance. Later in 2018, Khatami et al.  \cite{khatami2018sequential} proposed a two-step approach to shrink the search space. The proposed method used Radon projections as feature vectors for similarity comparison after narrowing the search-space by a convolutional neural network (CNN). To get a more meaningful Radon feature vector, the authors used the difference between two orthogonal projections for similarity search. This approach is validated on IRMA dataset, achieving an IRMA error of 168.05 (approximately 90.30\% accuracy), setting the benchmark on the dataset.

In early 2018, Tizhoosh and Babaie \cite{tizhoosh2018representing} introduced a new dense-sampling descriptor, based on Radon projections, called ``Encoded Local Projections'' (ELP). The authors build this histogram-based descriptor based on the highest gradient angle within each local neighborhood.  The angle of the highest gradient allows for capturing spatial projection patterns that are more descriptive and meaningful as compared to equi-distant angles. ELP is validated on three public datasets (IRMA, KIMIA Path24, and CT Emphysema), yielding a competitive accuracy measure against other established handcrafted descriptor, in several experimental settings. Later in 2018, Sharma et al. \cite{sharma2018facial} performed a study using ELP descriptor for facial recognition. In this setting, ELP was observed to perform better than LBP when used in the same configuration.

Presently, Deep Learning (DL) is the most trending research sub-field of Machine Learning (ML). 
The DL models are generally trained end-to-end, which greatly simplifies the training process. The popular architectures for image classification are Convolutional Neural Networks (CNNs). The first trainable CNN architecture was proposed by LeCun et al. in 1998~\cite{lecun1998gradient}. Almost a decade later, in 2012, AlexNet was developed at University of Toronto, establishing itself as the state-of-the-art model for image classification of that time~\cite{krizhevsky2012imagenet}. AlexNet achieved top-5 test error rate of 15.3\% on ImageNet classification challenge~\cite{deng2009imagenet}. The two major reasons for the success of AlexNet were, i)~availability of large amount of labelled data, and ii)~accelerated computing using GPUs. Since 2012, there has been a significant evolution of CNN architecture. The preference is given to deeper networks with smaller receptive fields, as the network becomes deeper its cumulative receptive field increases. An example of such a deep network is a 19-layer model often known as VGG-19 or OxfordNet that won the ImageNet challenge in 2014~\cite{simonyan2014very}. Furthermore, many improvements have been made in building blocks of CNNs architecture. These improvements include, inception module~\cite{szegedy2015going}---combines features from multi-resolution receptive fields at each layer, residual block~\cite{he2016deep}---uses residual learning for feature extraction, and dense block~\cite{huang2017densely}---extends the idea of residual learning to dense connections within layers. The performance of a deep model is highly dependent on availability of large amount of quality labeled data. The availability of large amount of labeled data is a limiting factor in medical image analysis due to shortage of experts, subjectivity of medical interpretations, and legal obligations to patients' privacy~\cite{razzak2018deep}. It is harder to adapt the performance of DL models in non-conventional problems (including medical domain), due to their black-box nature~\cite{miotto2017deep}. In healthcare, interpretability, quantitative performance and run-time performance of a ML technique are equally important. 

The motivation of this work is to employ Radon projections within shallow a neural topology to classify medical images. This should not only increase the interpretability of the network but also make the design, training and inference more practical and efficient.

\section{Projectron}
Projectron is a shallow neural network that uses Radon projections as input. We call the proposed architecture a ``Projectron'' because as a neural \emph{automaton} it uses \emph{projections}\footnote{The name \emph{projectron} has been used once in literature \cite{orabona2008projectron} but as it describes an algorithm and not a neural network, re-purposing the name for this work seems justified.}. Projectron, much like an MLP, is a supervised learning platform wherein the classification accuracy is dependent on the input features. A higher discrimination of features between each class usually results in a higher classification accuracy \cite{de2007feature}. In recent years, several descriptors have been introduced that have shown to complement learning techniques \cite{pawara2017comparing} \cite{driss2017comparison}; one such descriptor that has gained traction in the medical imaging domain are Radon projections \cite{sanz2013radon}.  

\begin{figure*}[htb]
	\centering
	\includegraphics[scale=0.45]{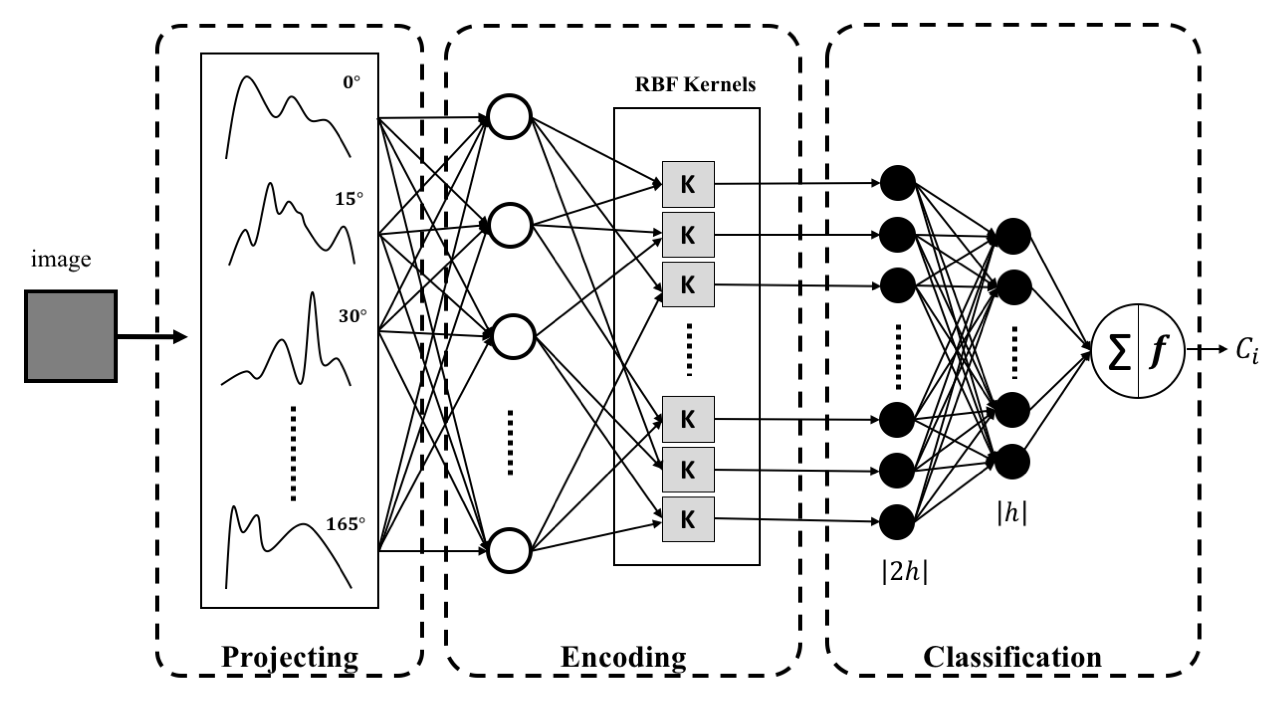}
	\caption{Projectron -- Classifying global projections. A small number of projection angles are used to generate parallel projections. The \emph{encoding stage} weights the projection values and pushes pairs of neuron outputs through RBF kernels. An MLP classifies the input image by using the output of the encoding stage.}
	\label{fig:radon_projection}
\end{figure*}

The proposed network takes in multiple \emph{global} Radon projections and pushes them through an encoding stage, and an MLP with two layers. Using global projections should contribute to increased interpretability as compared to ``local'' deep features. The Projectron learning is comprised of three phases: (\textit{i}) Applying Radon transform to provide parallel projections as image descriptors, (\textit{ii}) Encoding the projections using a one layer of neurons followed by a kernel layer using radial basis functions (RBFs), and (\textit{iii}) a shallow MLP with two layers for representation and classification. The following section will cover each of these phases in more detail. Fig. \ref{fig:radon_projection} provides a pictorial representation of the Projectron network.

\subsection{Projection Stage: Projecting the Image}
\begin{figure}[htb]
	\centering
	\includegraphics[width=0.9\columnwidth]{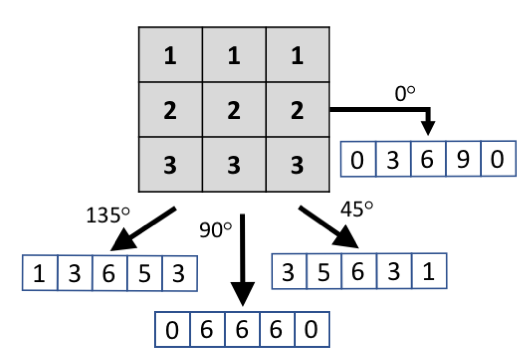}
	\caption{Computing Radon projections in a $3\times3$ window.}
	\label{fig:radon_sample}
\end{figure}

Radon transform, introduced by J. Radon in 1917 \cite{radon1986determination}, provides a profile that is a set of 1-dimensional projections of an object (Fig. \ref{fig:radon_projection}). The obtained projections of an object may be used to reconstructed the scene of the object at ${\rm I\!R}^n$ space, a technique called inverse Radon transform (i.e., filtered backprojection) \cite{toft1996radon}. Over the years, Radon transform has been adopted across various applications such as reconstructing images from computed axial tomography scans, and barcode scanners and computer vision \cite{toft1996radon}. Examining an image function $f(x, y)$, one can project $f(x, y)$ along a number of projection angles. Each Radon projection is essentially an integral transform of $f(x, y)$ which is a summation of values along lines constituted by each angle $\theta$ \cite{rey1990application}. These projections are used for assembling the \emph{sinogram} $R(\rho, \theta)$ with $\rho = x cos\theta + y sin\theta$. Hence, using the Dirac delta function $\delta(\cdot)$, the Radon transform of a two-dimensional image $f(x, y)$ can be defined as the integral along a line inclined at an angle $\theta$ and at a distance $\rho$ from the origin \cite{seo2004robust} (see Fig. \ref{fig:radon_sample}):
\begin{equation}
R(\rho, \theta) = \int\limits_{-\infty}^{\infty} \int\limits_{-\infty}^{\infty} f(x, y)\delta(x cos\theta + y sin\theta - \rho) dx dy,
\end{equation}
\noindent where $-\infty < \rho < \infty, 0 \leq \theta < \pi$. 

For the Projectron, Radon projections are computed for every image from $0^{\circ}$ to $180^{\circ}$, with an $\theta = 15^{\circ}$, empirically chosen, to provide a total of $12$ projections per image. Each image is gray-scaled and re-sized to have the same dimensions such that the projection length across all angles are the same (see Algorithm \ref{alg:preprocess_algo}).

\begin{algorithm}[htb]
\KwIn{Query image \textit{$I$}, Image dimension \textit{$d \times d$}, Projection angles \textit{$\theta$}}
\KwOut{Radon image descriptor \textit{$I_f$} which consists of number of projections angles \textit{$N$} and projection length \textit{$feat$}}

\SetAlgoLined\DontPrintSemicolon
\SetKwFunction{proc}{getFeatures($I,d \times d, \theta$)}
\SetKwProg{myproc}{Procedure}{}{}
\myproc{\proc}{
\nl $I_{p} \gets Resize(I, (d,d))$ (re-sizes $I$ to $d \times d$)\;
\nl $I_{p} \gets Grayscale(I_{p})$ (convert $I_{p}$ to grayscale)\;
\nl $I_{f} \gets Radon(I_q, \theta), axis=1)$ (get Radon projection per image of shape ($N$,$feat$) for each $\theta$ \textcolor{blue}{angle which is acquired between two consecutive projections\;
\nl \KwRet{$I_{f}$} \;}}

\caption{{\bf Projecting images} 
\label{alg:preprocess_algo}}
\end{algorithm}

\subsection{Encoding Stage: Kernelizing Weighted Projections}
In simple terms, the encoding block is responsible for learning the relationship between the projections using a layer of neurons followed by a kernel layer. After extracting the projections for each equi-distant angle $\theta$, these projections are provided to a first layer of neurons which encodes the inputs by adjusting the synaptic weights $\vec{w}\in{\rm I\!R}^n$, $b\in{\rm I\!R}$ bias, and $\varphi:  \rm I\!R \rightarrow \rm I\!R$ is an activation function. Hence, given an input vector $\vec{x}\in{\rm I\!R}^n$, each transformation is defined by $\varphi(\langle\vec{w}, \vec{x}\rangle + b)$.The weights $\vec{w}$ and bias $b$ are generally adjusted based on the \emph{ReLu} activation function \cite{nair2010rectified}. Introduced in 2000 by Hahnloser et al.  \cite{hahnloser2000digital}\cite{hahnloser2001permitted}, the Rectified Linear Unit (ReLu) is an established activation function that is found to accelerate the convergence of stochastic gradient descent algorithm when compared to \emph{sigmoid} and \emph{tanh} functions \cite{krizhevsky2012imagenet}. This activation function is simply thresholded to zero wherein the function outputs either $0$ or a positive value for every $x$ input to a neuron: $f(x) = max(0, x)$.

The outputs of the first layer in the encoding stage is passed onto a layer with RBF kernels to enforce an increase in linear separability as we intend to keep the Projectron rather shallow. To better capture the variations of each projection, we use the Gaussian basis function to center each vector and set the $\gamma$ value as a dynamic variable. Hence, the variable $\gamma$ is weighted to minimize the overall classification error. The RBF takes two parameters as inputs that determines the center (mean) value of the function to provide a desired output value. The RBF is a real-valued function whose value depends only on the distance from the origin, so that $\phi(\mathbf{x}) = e^{-(\gamma \mathbf{x})^2}$. For our purposes, the RBF value depends on the distance from some other point $x_i$, so that
\begin{equation}
\phi(\mathbf{x}, \mathbf{x_i}) = e^{-(\gamma \lVert\mathbf{x}-\mathbf{x_i}\rVert)^2}.
\end{equation}
The sums of RBFs are typically used to approximate given functions. This approximation process can also be interpreted as a simple type of neural network that we placed after the first layer. The RBF approximations are of the form 
\begin{equation}
y(x) = \sum_{i=1}^{N} w_i \phi (\lVert \mathbf{x}-\mathbf{x_i} \rVert),   
\end{equation}
\noindent where $\gamma\in[0,10]$ is a dynamic variable that we learn. The approximating function $y(x)$ is the sum of $N$ RBFs, each of which are associated with a different center $x_i$, and weighted by coefficient $w_i$. To learn the gamma variable $\gamma$, every other encoding neuron output is provided as an input pair to the RBF kernel. The RBF layer computes the distance between the inputs to enhance the discrimination to the classification block that follows. 

\subsection{Classification Stage: Shallow MLP}
The RBF layer is connected to a shallow Multi-Layer Perceptron (MLP) \cite{el2013extraction} to perform classification. The MLP is among the most useful types of neural networks, with an ability to learn the representation of data and relate it to the output, increasing the overall accuracy. In this case, the MLP is structured to have an input layer to be the RBF layer output from encoding stage, followed by a single hidden-layer with ReLU activation functions. The last layer is the classification layer which is based on softmax function to calculate the probability distributions. In particular, the softmax function derives  probabilities for each class and for every image. The class that has the highest probability (i.e., closer to 1) is considered the predicted label. The weights of the shallow MLP may as well be employed as a ``represnetation'' for the input image for other purposes (e.g., image search). All steps to train the Projectron are described in Algorithm \ref{alg:projectron_algo}.
\begin{algorithm}[t]
\small
\KwIn{Radon descriptor \textit{$I_{f}$} of shape \textit{($N, feat$)}, epochs \textit{$epoch$}, Batch Size \textit{$batchSize$}, true labels per image \textit{$labels$}}
\KwOut{Classification label \textit{$prediction$}}

\SetAlgoLined\DontPrintSemicolon
\SetKwFunction{proc}{Projectron($I_{f}, epoch, batchSize, labels$)}
\SetKwProg{myproc}{Procedure}{}{}
\myproc{\proc}{
\nl $I_{d} \gets Flatten(I_{f})$ (flatten Radon descriptor to 1-D vector) \;
\nl $numClass \gets max(labels)$ (get total number of classes) \;

\tcc{Define placeholders for perceptron}
\nl $inFeat \gets Placeholder(None, I_{d})$ (define input features as placeholder) \;
\nl $inClass \gets Placeholder(None, numClass)$ (define classes as placeholder) \;

\tcc{First learning layer, the perceptron}
\nl $layer \gets Dense(inFeat, activation=``ReLu'')(inFeat)$ (Perceptron layer with ReLu activation function) \;

\tcc{Define gamma for RBF distribution}
\nl $\gamma \gets Variable(Random(length(N-2)\times 10))$ (declare $\gamma$ as dynamic variable for every other projection) \;
\nl $layer \gets RBF(layer, \gamma)$ (provide the perceptron output to RBF layer as input along with $\gamma$) \;

\tcc{Multi-layer perceptron}
\nl $layer \gets Dense(layer, activation=``ReLu'')(layer)$ (provide RBF output as input to MLP) \;
\nl $layer \gets Dense(layer/2, activation=``ReLu'')(layer)$ (compress input layer to half its dimension) \;
\nl $classify \gets Dense(numClass, activation=``linear'')(layer)$ (compress previous layer onto number of classes) \;

\tcc{Define error and predictions}
\nl $error \gets Softmax(labels, classify)$ (get error by comparing the predicted label to ground-truth using softmax classification) \;
\nl $predictions \gets max(classify)$ (get predicted outputs) \;
\nl \KwRet{$predictions$} \;}

\caption{{\bf Projectron: Learning projections} 
\label{alg:projectron_algo}}
\end{algorithm}
\section{Experiments}
A total of five different experiments were conducted on publicly available datasets to evaluate Projectron against MLP with raw images and Radon features as input. These datasets include a non-medical dataset (MNIST) and four medical datasets, namely CT Emphysema, Invasive Ductal Carcinoma (IDC), IRMA, and Pneumonia. The following section briefly describes these datasets. Thereafter, the classification results of Projectron against the MLP are reported for each dataset. 

\subsection{Datasets} \label{datasets}
\subsubsection{MNIST Dataset} 
The MNIST dataset \cite{lecun1998mnist} is among the most popular image processing multi-class dataset and is comprised of several thousands of handwritten digits. In particular, there are a total of 70,000 images depicting digits 0 to 9, which are first pre-processed using min-max normalization. The dataset is pre-distributed with 60,000 images for training and 10,000 images are for testing. For training, each image is processed at its original resolution sized at $28\times28$. 
\subsubsection{Emphysema Dataset} 
For this study, we also used the ``Computed Tomography Emphysema'' dataset \cite{sorensen2010quantitative} which contains 168 CT  patches of size $61\times61$ from 115 high-resolution CT slices. These scans are gathered from 39 patients, and each patch is manually annotated into one of three categories: (\textit{i}) 59 observation of normal tissue, (\textit{ii}) 50 observations of Centrilobular Emphysema (CLE), and (\textit{iii}) 59 observations of Paraseptal Emphysema (PSE). For training, the images were re-sized to $62\times62$ prior to training and testing using leave-one-out approach - i.e. a total of 168 models were trained. The accuracy is measured based on the cardinality of correctly classified images $C$. The accuracy $A_{CT}$ can be calculated as
\begin{equation} \label{emphysema}
A_{CT} = \frac{|C|}{168}.
\end{equation}
\subsubsection{IDC Dataset} 
Invasive Ductal Carcinoma (IDC) is the most common subtype of all breast cancers detected in histopathology slides. To grade the whole slide image (WSI), pathologists typically focus on the IDC region. The dataset is retrieved from Kaggle website, and consists of 162 WSIs in total \cite{cruz2014automatic} \cite{janowczyk2016deep}. Slides are from the Hospital of the University of Pennsylvania and The Cancer Institute of New Jersey. All slides were digitized and scanned at 40x magnification. Each slide is broken down into 277,524 patches of size $50\times50$ with 198,738 being IDC negative and 78,786 diagnosed as IDC positive. The training and testing were distributed to 114,235 and 50,963 instances, respectively. For training, each image was re-sized to $50\times50$ and converted into gray scales. Similar to emphysema, the correctly classified images $C$ is compared against the test set for total accuracy measure as follows:
\begin{equation} \label{idc_acc}
A_{IDC} = \frac{|C|}{50963}.
\end{equation}
\subsubsection{IRMA Dataset} 
IRMA is a retrieval dataset of radiography images. This x-ray dataset is comprised of 12,677 training and 1,733 testing images created from clinical cases at the Department of Diagnostic Radiology at RWTH Aachen University. Each image is annotated using an IRMA code which is comprised of four mono-hierarchical axes: the technical code (T) for imaging modality, directional code (D) for body orientations, anatomical code (A) for the body region being imaged, and biological code (B) for the biological system examined. The IRMA code is 13 characters in length of form: TTTT-DDD-AAA-BBB, wherein each can range from {0, 1,..., 9; a, b,...,z} \cite{lehmann2003irma}. The IRMA code is evaluated when comparing the IRMA codes between the retrieved and the ground-truth. The IRMA error is defined as \cite{mller2010image}
\begin{equation} \label{irma_error}
error = \sum_{i=1}^{n_{\textrm{char}}} \frac{1}{b_i} \frac{1}{i} g(l_i, \hat{l}_i),
\end{equation}
\noindent where $l_i$ is the IRMA code of the query image, $\hat{l}_i$ the IRMA code of the retrieved image, $b_i$ the number of possible states for each position, $n_char$ is the number of characters on the axis, and $g(\cdot)\in[0,1]$ is a function for correct/wrong matching. The total error is then defined as follows
\begin{equation} \label{total_irma}
A_{IRMA} = 1-\frac{1}{1733}\sum_{i=1}^{n_{\textrm{char}}} \frac{1}{b_i} \frac{1}{i} g(l_i, \hat{l}_i).
\end{equation}
For training, each image is re-sized to $147\times147$. 
\subsubsection{Pneumonia Dataset} 
Pneumonia is an infection of the lungs which results in an inflammation in the air sacs making it difficult for the patient to breathe. This is a dataset retrieved from Kaggle which consists of 5,863 x-ray images classified either as ``pneumonia'' or ``normal'' \cite{kermany2018identifying}. These chest x-rays were selected from pediatric patients between \textcolor{black}{1 to 5 years of age from Guangzhou Women and Children's Medical Center}. Since each image is sized differently in this dataset, we re-sized each image to $150\times150$. As for the distribution of the dataset - 70\% was randomly selected for training, and the remaining 30\% of the data was selected for testing the Projectron and MLP.

\subsection{Parameter Setting}
The implementation of the Projectron is straight-forward. The first step is to gray-scale (if necessary) and re-size all images. For each image, the projection gap at $\theta=15^{\circ}$ is empirically chosen - generating a total of 12 projections per image: $\{0^\circ, 15^\circ,30^\circ,\dots,165^\circ\}$. The length of the projection is generally equal to the hypotenuse length of the image with zero padding when necessary. The Radon features for each image is provided to the Projectron which classifies images based on ReLU activation using Adam optimizer. Since Radon features are global representation of the image, the run time for Projectron is a lot quicker as compared to MLP with raw images as input. To avoid overfitting, the accuracy and loss is calculated per epoch, and the training is terminated when the loss per epoch stays the same or drops for every iteration. For sake of comparison, we also train and test (conventional) MLPs with raw images as input, the re-sized images that were computed for the Projectron are also used as input to the MLP. In terms of architecture, the MLP is a shallow network, with either one or two hidden layers followed by a softmax classification. Not only does the MLP with raw images takes a longer time to train, it also provides a comparable or even lower accuracy for most of the reported datasets. Finally, the use of MLP with Radon feature as input was also tested and observed to be a competitor to the Projectron. This, of course, empirically confirms one of our assumptions, namely that using projections instead of raw data is useful when shallow architectures are preferred. In this case, the inputs to the MLP is the exact same Radon features that were provided to the Projectron. This technique resulted in a better accuracy when compared to MLP with raw images. \textcolor{black}{However, Projectron yields a better result on the IDC, IRMA, and Pneumonia datasets when compared against both of the aforementioned approaches. As for the MNIST and CT Emphysema dataset, Projectron yields a lower-yet-comparable accuracy.}

\subsection{Results}
Table \ref{tab:final_results} shows the results for all five datasets, wherein the reported values are testing accuracy and the total number of trainable parameters $\lvert h \rvert$. For all datasets, the images were gray-scaled and re-sized. In addition, early-stopping with a patience of three epochs is adopted to avoid overfitting for each approach across all datasets. 

It was observed that MLP with raw images trains the slowest with Projectron training as quick as MLP with Radon projections even though it has the highest number of trainable parameters. For instance, in the MNIST dataset, each image is re-sized to $28 \times 28$, yielding a total trainable parameters of $1,196,100$ for Projectron which took roughly 2.5 minutes to train. This is similar to MLP with Radon projections, which had $117,850$ parameters and took a little more than 2 minutes to train . In comparison, a one-hidden layer MLP with raw images has $311,650$ trainable parameters and took more than 4 minutes to train. 

For a better comparison, we constructed a \textbf{deep MLP} with 7 hidden layers to increase the total number of parameters to be comparable to that of Projectron. Keeping the reduction of each layer the same (i.e., half the length of the input), the deep MLP had $1,156,998$ trainable parameters, achieving a similar testing accuracy when compared to its shallow MLP. For testing accuracy, please refer to Table \ref{tab:final_results}. 

\begin{figure*}[htb]
	\centering
	\includegraphics[width=0.6\textwidth]{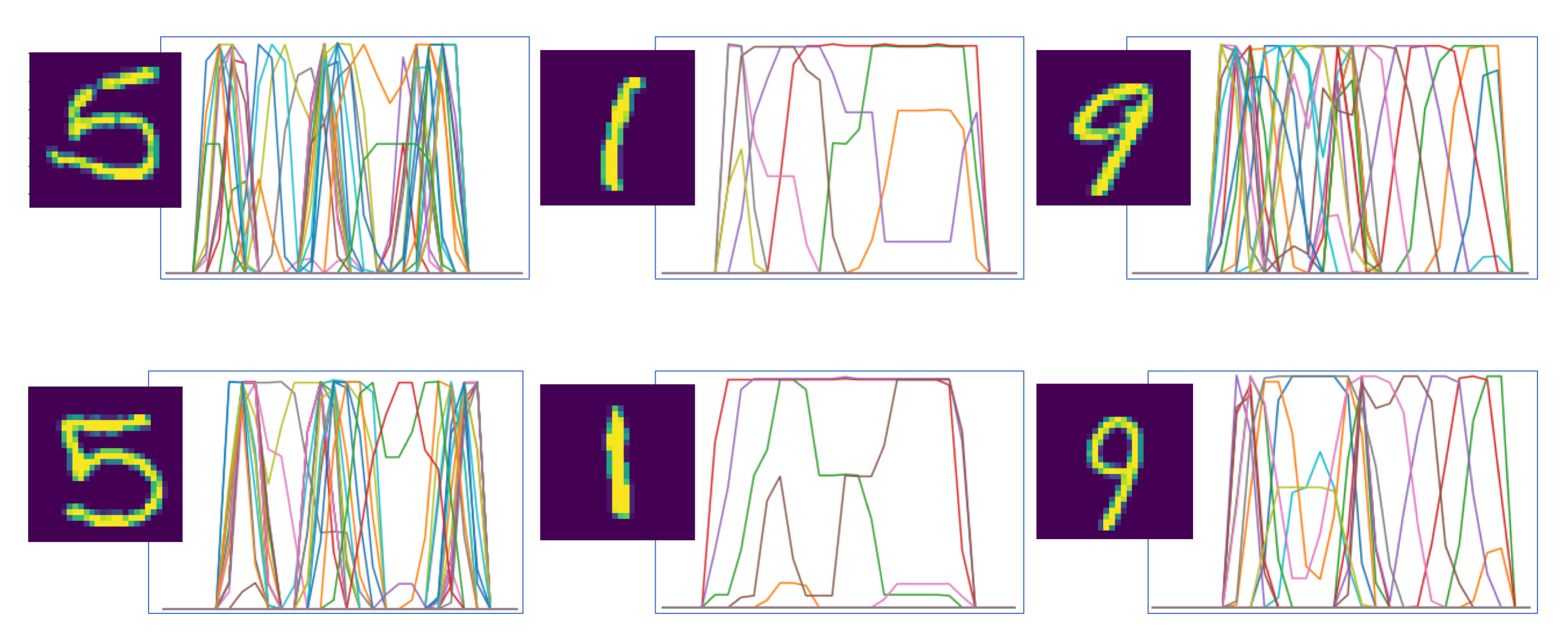}
	\caption{Comparing projections for classified digits can shed light on the \emph{reason for classification}.The global nature of projections and their small numbers enables us to understand the rational for classification.}
	\label{fig:radon_interpret1}
\end{figure*}

\begin{figure*}[htb]
	\centering
	\includegraphics[width=0.8\textwidth]{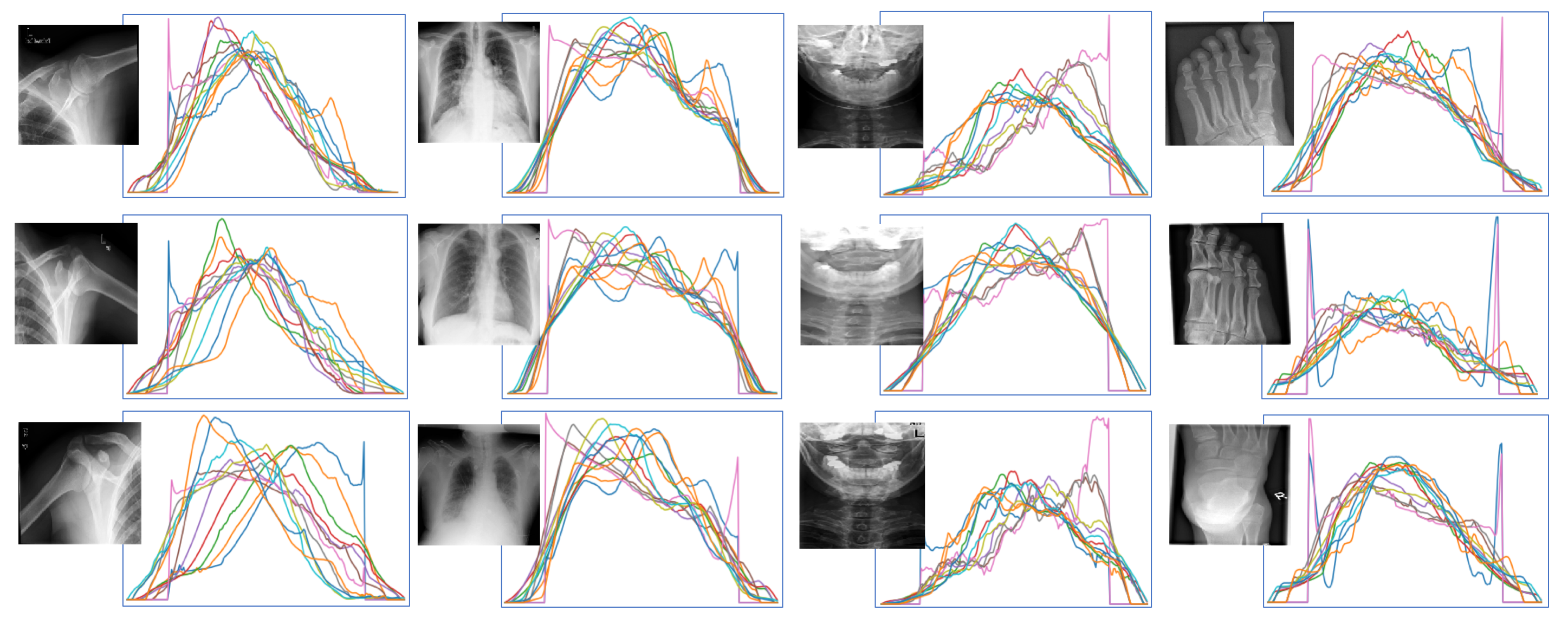}
	\caption{Global projections (the inputs to the Projectron) are well-understood in the medical imaging. The mixture and typicality of these projections, their shape and relationships to the image content can be easily interpreted by the human expert. For instance, for the lung x-ray (2nd column from left), the two valleys of the blue projections represent the lungs whereas projections at $30^\circ$ and $150^\circ$, as well as projections at $45^\circ$ and $135^\circ$ are extremely similar as they both go across both lungs, respectively. Such relationships can be visualized more apparently to assist the human expert in interpreting the results.}
	\label{fig:radon_interpret2}
\end{figure*}

Overall each strategy was able to generalize the dataset well. This experiment shows that a shallow network, such as Projectron, can generalize and learn the features just as good as a deeper network. Moreover, the global Radon features extracted in Projectron network is more interpretable than deep local features as observed when comparing classes in the IRMA dataset in Figs. \ref{fig:radon_interpret1} and \ref{fig:radon_interpret2}.

The Projectron is observed to outperform both the MLP approaches for the IDC, IRMA and Pneumonia datasets. For these datasets, each image is re-sized to $50\times50$, $147\times147$, and $150\times150$ respectively. The MLPs are observed to struggle when learning the image features, forming a bias towards one of the classes during classification. This could be due to the variation of images in the dataset as well as noise present in the images. As for the relatively smaller dataset, CT Emphysema, leave-one-out-approach was adopted wherein the images is re-sized to $62\times62$ for training. In this dataset, MLP with Radon achieves the best result at $60.12\%$, followed by Projectron at $59\%$ and MLP with raw images at $57.73\%$. For this dataset, each network is able to classify between the three classes well. 

Generally, the Projectron achieves competitive accuracy results if not better than convetional MLPs for multiple datasets. Not only is Projectron a shallow and more explainable neural network, it is also observed to train faster than MLP with raw images as input, and learn the images better in most cases. 

\begin{table}[htb]
\normalsize
\caption{Results for Projectron against two shallow MLP networks with raw images and Radon projections as input. Results include the trainable parameters ($\lvert h \rvert$) for each approach.}
\begin{tabular}{lll}
\hline
\hline
\multicolumn{3}{c}{\textbf{MNIST Dataset}}                                                                     \\ \hline
\multicolumn{3}{c}{(60,000 images training and 10,000 images for testing)}                                     \\ \hline
Method                     & Accuracy                     & $\lvert h \rvert$                     \\ \hline
MLP+Raw                    & 97.89\%                      & 311,650                        \\
MLP+Radon                  & 96.81\%                      & 117,850                         \\ 
Projectron                 & 94.75\%                      & 1,196,100                       \\\hline
\multicolumn{3}{c}{\textbf{CT Emphysema Dataset}}                                                              \\ \hline
\multicolumn{3}{c}{(168 images for training/testing)}                                                          \\ \hline
Method                     & Accuracy                     & $\lvert h \rvert$                     \\ \hline
MLP+Radon                  & 60.12\%                             & 560,212                \\
Projectron                 & 59.00\%                             & 6,884,958                       \\ 
MLP+Raw                    & 57.73\%                             & 7,397,782                        \\ \hline
\multicolumn{3}{c}{\textbf{Invasive Ductal Carcinoma}}                                                         \\ \hline
\multicolumn{3}{l}{(114,235 images for training and 50,963 images for testing)}                                \\ \hline
Method                     & Accuracy                     & $\lvert h \rvert$                     \\ \hline
Projectron                 &  78.00\%                            & 3,754,918                        \\
MLP+Raw                    & 71.35\%                             & 3,128,752      \\
MLP+Radon                  & 71.35\%                             & 364,232                        \\ \hline
\multicolumn{3}{c}{\textbf{IRMA Dataset}}                                                                      \\ \hline
\multicolumn{3}{l}{(12,677 images for training and 1,733 images for testing)}                                  \\ \hline
Method                     & Accuracy                     & $\lvert h \rvert$                     \\ \hline
Projectron                 & 69.00\%                             & 38,548,291                        \\
MLP+Radon                  & 47.02\%                             & 3,187,449                        \\
MLP+Raw                    & 41.23\%                             & 233,582,490                     \\ \hline
\multicolumn{3}{c}{\textbf{Pneumonia Dataset}}                                                                 \\ \hline
\multicolumn{3}{l}{(5,216 images for training and 624 images for testing)}                                     \\ \hline
Method                     & Accuracy                     & $\lvert h \rvert$                     \\ \hline
Projectron                 & 70.03\%                             & 33,767,754                        \\
MLP+Raw                    & 37.50\%                             & 253,158,752                         \\
MLP+Radon                  & 37.50\%                              & 3,270,404                         \\ 
\hline
\hline
\end{tabular}
\label{tab:final_results}
\end{table}

\section{Summary and Conclusions}
In this paper, a new artificial neural network called ``Projectron'' was introduced. The Projectron learns a small number of Radon projections captured from images. The proposed network is comprised of three phases, namely acquiring projections, encoding projections, and classifying the results with a shallow MLP. For each image, a small number of projections are obtained with an equi-distant angle of $\Delta=15^{\circ}$ between $\theta=0^{\circ}$ and $\theta=180^{\circ}$. These projections are provided to a layer of neurons to weight them. The output is then forwarded to a  layer of RBF kernels which is supposed to increase the linear separability between each pair of weighted projection values. Finally, a shallow MLP with only two layers is adopted to classify the encoded features using ReLu activation for the hidden layer, and ``Softmax'' activation on the output layer. We validated the Projectron on five public datasets. The Projectron was observed to perform better than MLP approach for IDC, IRMA and Pneumonia datasets. For the MNIST and Emphysema dataset, the Projectron performs competitively with MLP approaches. For all the dataset, Projectron learns the dataset much quicker than a traditional MLP with an input of either raw images or Radon projections. Finally, the Projectron seems to generalize the dataset better -- which is apparent when examining the loss per epoch graph as well as classification results. \textcolor{black}{For future work, we would like to explore the possibilities of using Radon projections for extracting features for a deeper network - a Convolution Neural Network inspired architecture. Also, we would like to further improve the Radon features by  determining an optimal angle $\theta$ at which the projection contains the most relevant features.}

The proposed Projectron demonstrates the potential for classifying medical images on a par with or even better than established MLP networks. The \emph{shallowness} of Projectron is certainly beneficial for fast  developments. As well, incorporating global projections does in fact increase the \emph{interpretability} of the classification for medical images.

\bibliographystyle{splncs}
\bibliography{references}

\end{document}